\documentclass[11pt]{article}
\usepackage[left=2.7cm,bottom=2.7cm,right=2.7cm,top=2.7cm]{geometry}
\usepackage[square,numbers]{natbib} 
\usepackage{booktabs} 
\usepackage{color}

\usepackage[utf8]{inputenc} 
\usepackage[T1]{fontenc}    
\usepackage{hyperref}       
\usepackage{url}            
\usepackage{booktabs}       
\usepackage{amsfonts}       
\usepackage{nicefrac}       
\usepackage{microtype}      
\newif\ifworkinprogress
\workinprogresstrue
\ifworkinprogress

  \newcommand{\pv}[1]{\textcolor{blue}{\small{[Puya] #1}}}
  \newcommand{\RM}[1]{\textcolor{red}{\small{[RM] #1}}}
\else
  \newcommand{\mq}[1]{}
  \newcommand{\es}[1]{}
  \newcommand{\mr}[1]{}
  \newcommand{\ty}[1]{}
  \newcommand{\pv}[1]{}
\fi

\usepackage{csquotes}
\usepackage{epsfig}
\usepackage{color}

\usepackage{balance}  
\usepackage{algorithmic}
\usepackage{algorithm}
\usepackage{multirow}
\usepackage{graphicx}
\usepackage{amsmath}
\usepackage{amssymb}
\usepackage{amsfonts}
\usepackage{xspace}
\usepackage{url}
\usepackage{./tweaklist}
\usepackage{hyperref}
\usepackage{booktabs}
\usepackage{bbold}
\newcommand{\normaltilde}{\raise.17ex\hbox{$\scriptstyle\sim$}}

\newcommand{\IGNORE}[1]{}

\newcommand{\pandora}{{\sf PMusic}}

\newcommand{\lastfm}{{\sf lastfm}}
\newcommand{\MAE}{{\sf MAE}}
\newcommand{\XGBoost}{{\sf XGBoost}}
\newcommand{\Baseline}{{\sf Baseline}}

\newcommand{\BaselineSIGIR}{{\sf SIGIR2017}}

\newcommand{\ModelOne}{{\sf Model1}}
\newcommand{\Ridge}{{\sf Ridge}}
\newcommand{\ModelTwoLTwo}{{\sf Model2-L2}}
\newcommand{\ModelTwoLOne}{{\sf Model2-L1}}
\newcommand{\ModelTwoGBT}{{\sf Model2-GBT}}
\newcommand{\ModelThreeLTwo}{{\sf Model3-L2}}

\newcommand{\ModelThreeGBT}{{\sf Model3-GBT}}

\newcommand{\commentedtext}[1]{}

\newtheorem{theorem}{Theorem}

\newtheorem{proof}{Proof}
\newcommand{\B}{\boldsymbol}
\DeclareMathOperator*{\argmin}{argmin}

\DeclareMathOperator*{\sign}{sign}

\graphicspath{{./figures/}}

\usepackage{authblk}
\usepackage{natbib}
\begin{document}
\title{Hierarchical Modeling and Shrinkage for User Session Length Prediction in Media Streaming}


\author[1]{Antoine Dedieu\thanks{adedieu@mit.edu}}
\author[1]{Rahul Mazumder\thanks{rahulmaz@mit.edu}}
\author[2,3]{Zhen Zhu\thanks{zzhu@pandora.com}}
\author[2]{Hossein Vahabi\thanks{puya@pandora.com}}
\affil[1]{Massachusetts Institute of Technology}
\affil[2]{Pandora Media, Inc.}
\affil[3]{Stanford}

\maketitle

\begin{abstract}
An important metric of users' satisfaction and engagement within on-line streaming services is the \emph{user session length}, i.e. the amount of time they spend on a service continuously without interruption. Being able to predict this value directly benefits the recommendation and ad pacing contexts in music and video streaming services. Recent research
has shown that predicting the exact amount of time spent is highly nontrivial due to many external factors for which a user can end a session, and the lack of predictive covariates. Most of the other related literature on duration based user engagement has focused on dwell time for websites, for search and display ads, mainly for post-click satisfaction prediction or ad ranking.

In this work we present a novel framework inspired by hierarchical Bayesian modeling to predict, at the moment of login,  the amount of time a user will spend in the streaming service. The time spent by a user on a platform depends upon user-specific latent variables which are learned via hierarchical shrinkage. Our framework enjoys theoretical guarantees and  naturally incorporates flexible parametric/nonparametric models on the covariates, including models robust to outliers. Our proposal is found to outperform state-of- the-art estimators in terms of efficiency and predictive performance on real world public and private datasets.


\end{abstract}

%
%

\section{Introduction}
\label{sec:introduction}
%


On-line streaming services such as Pandora, Netflix, and Youtube constantly seek to increase their share of on-line attention by keeping their users as engaged as possible with their own services \cite{Mounia2016}. A well known challenge is how to measure the users' engagement, and what are the key components that create an engaging streaming service. One important engagement metric is the amount of time spent by users within the service. When users access streaming services, they usually watch videos, movies, on-line TV or listen to music, and after a while they leave the service.  
We refer to the user interaction from the moment they start the service to the moment they leave as a \textit{user session}, and the time spent during one session as \textit{user session length} \cite{SIGIR2017}.

In this paper, we aim to predict the user session length using real world datasets from two music streaming platforms, i.e. predicting, at the beginning of the session, the amount of time they will spend listening music. 
Understanding and modeling the factors that can affect the session length is of great use for various downstream tasks.
In fact it allows recommender systems to tune the explore vs exploit parameters for each user. In addition, having an accurate estimate of the users' session lengths allows the streaming service to adjust ad pacing per user. Ads can be rescheduled in a way to keep the revenue target (i.e. total number of ads presented) as well as improve user experience. 

Predicting the length of user sessions is very challenging as the authors reported in \cite{SIGIR2017}. First, user sessions can end for many different external reasons that have nothing to do with quality of the streaming services, such as moving to subway, reaching home, and most of these contextual covariates are not easily accessible because of technological or privacy reasons \cite{SIGIR2017}. Second, a certain amount of users are casual users in the sense that they only use the streaming services a few times per month, which makes the problem of estimating session lengths for those users hard. 
Furthermore, a lack of predictive covariates makes it harder to correctly predict what is going to be the next session length for a user.

A first approach toward session length analysis and prediction \cite{SIGIR2017} is based on a Boosting algorithm. Most of the other related research were focused on modeling the time spent after clicking a search result \cite{borislov2016,Microsoft2014satisfaction} or advertisement \cite{Barbieri2016}. The best models are based on survival analysis, mainly because data are censored in these applications. In fact, after clicking of a search result or an ad, users can either turn back to the search page or can abandon the final page, instead of turning back to the main search page. This is not the case of session length data that is not censored\footnote{Observations are called censored when the information about their survival time is incomplete.} -- this opens the door to using a suite of methods based on regression modeling, shrinkage and relevant generalizations. Furthermore, in the case of web search or ad click, the user enters a query or click and checks the results, therefore the intersection between search or ad title and the landing page give highly predictive features \cite{Microsoft2014satisfaction}. In the case of a streaming service, the user interaction can be very low, but still they can have very long sessions ("lean-back" behavior)~\cite{SIGIR2017}.        

In this paper, we propose a novel framework inspired by hierarchical Bayesian modeling and shrinkage principles that allows us to express session lengths in terms of user-specific 
latent variables. These variables are estimated via a joint learning framework which is rather broad in scope -- we use Bayes, Empirical Bayes and MAP estimation techniques--particular choices are based on computational tractability considerations.
Informally speaking, our models learn by borrowing strengths across all users and   
making use of rich covariate information. We also propose
models that can incorporate outliers in data.
A salient aspect of our framework is its modularity -- it includes state-of-the-art models as special cases, and naturally allows for a hierarchy of flexible generalizations. Hence, it allows the practitioner to glean insights about the problem, by assessing incremental gains in predictive accuracy associated with different generalizations and the incorporation of covariate-information. 
We present tailored algorithmic approaches based on modern large convex optimization techniques to address the computational challenges.
We summarize some of our key findings in this paper:
\begin{itemize}
	\item We outperform a baseline estimator by a margin of $13\%$ to $19\%$ and the state-of-the art in session length prediction \cite{SIGIR2017} up to $4.3\%$ in terms of \emph{Mean Absolute Error} measured in seconds on $2$ different real world datasets. 
    \item We show that some of our proposed prediction models can be (a) more accurate than state of the art (with 1-2\% relative improvement in prediction performance), (b) between $22$ to $43$ times faster in training time; and (c) $30$-$500$ times faster in prediction time, reaching around 1ms.
    \item We provide a modular framework specifically for this problem that allow more flexible generalizations, 
\end{itemize}


\section{Key idea}
\label{sec:probdef}
%
The main focus of this paper is the prediction of user session length at the moment of login into a streaming service. For this purpose, we exploit the users past interaction with streaming services, previous sessions lengths, build a set of features (as we will describe in Section ~\ref{sec:datasets}), and we set a learning framework for prediction purposes. We proceed by using a Gaussian approximation for the distribution of the 
log of a users' session length (Section ~\ref{sec:datasets}). As we will see, this will help us in creating a well-grounded and tractable inferential framework. Since we want a prediction framework that is performing well for very active users (with many sessions) and less-frequent users (with only a few sessions per month), we propose a formal framework inspired by hierarchical Bayesian shrinkage ideas that allows us to model the session-length of a user in terms of user-specific latent variables.
Fundamental principles of Bayesian shrinkage and estimation encourage users to borrow strengths across each other, including (but not limited to) covariate information. This (i) ameliorates the high variance (and hence low predictive accuracy) of user-specific maximum likelihood estimates for less-frequent users; and (ii) leads to an overall boost in prediction accuracy for more frequent users as well.  
Bayesian decision theory and 
empirical Bayes methodology~\cite{efron2012large}
provides a formal justification of our framework.
The notion of shrinkage that we undertake here is quite broad: It applies to cases with or without covariate information; by a flexible choice of priors on the latent variables, we can also build models robust to heavy tailed errors. We show that the state-of-the-art model~\cite{SIGIR2017} on this particular problem is a special case of our framework.  For flexible models and/or priors when Bayes estimators become computationally demanding, we recommend MAP estimation for computational efficiency, this includes the case for robust modeling with a Huber loss~\cite{huber2011robust}. We resort to techniques in modern large scale convex optimization to achieve computational scalability and efficiency.


\section{Mathematical Framework}
\label{sec:model}
%
We formally develop the inferential framework to address the session length prediction problem. We denote the total number of users by $N$. For every
$i \in [N]:=\{1, \ldots, N\}$, let $n_i$ be the number of past sessions of user $i$, and $\tilde{y}_{ij}$ be the time spent by user $i$ in the $j$th session. We will work with 
log of session length $y_{ij}=\log(\tilde{y}_{ij})$ as our response as this gives a better approximation to the Gaussian distribution (See Table~\ref{table:pandora-norm-stats}).
We note that similar (variance stabilizing) transformations\footnote{We note that it is also possible to fine-tune the transformation by considering a family of the form: $\log(\cdot)^{\tau}$ for $\tau>0$; and optimizing over $\tau$ to obtain the closest approximation to a Gaussian distribution in terms of the Kolmogorov-Smirnov goodness of fit measure (for example). However, we do not pursue this approach here; as it leads to marginal gains in eventual prediction performance.} are often used in the empirical Bayes literature~\cite{efron2012large} so that one can take recourse to the rich literature in Gaussian estimation theory.
We denote by  ${\B{y}}_i = \left({y}_{ij}\right)_{j \in  [n_i] }  \in \mathbb{R}^{n_i}$ a vector of 
log-session-lengths of user $i$; $N_0 = \sum_{i=1}^N n_i$ the total number of sessions across all users; and ${\B{y}} = ({\B{y}}_i)_{i \in [N]} \in \mathbb{R}^{N_0}$ is a vector of log-session-lengths of all users across all sessions.
In addition, covariate information per-session is available (See Section~\ref{sec:datasets}). For a given user, some of these features are fixed across sessions (age, gender...) and others depend upon the session (network, device...).
Let $\B{x}_{ij} \in \mathbb{R}^d$ denote covariates corresponding to the $j$th session of user $i$.  $\B{X} \in \mathbb{R}^{N_0\times d}$ denotes a matrix with rows $\B{x}_{ij}$ that are stacked on top of each other. The $k$th row of $\B{X}$ corresponds to $k$th entry of the response vector ${\B{y}}$. The columns of $\B{X}$ were all mean-centered and standardized to have unit $\ell_{2}$ norm.


A natural point estimate of the time spent by the $i$th user may be taken to be the average time spent by the user in past sessions -- however, we see that this does not lead to good predictions due to the high variance associated with users with few sessions. This behavior is not surprising
and occurs even in the simple Gaussian sequence model 
as explained in Section~\ref{sec:bayesian}.

\subsection{Review of Bayes, Empirical Bayes and MAP}\label{sec:bayesian}
This section provides a brief review of Bayes, Empirical Bayes (EB) and Maximum A posteriori (MAP) estimation in the context of the Gaussian sequence model~\cite{efron2012large}. The exposition in this section lays the foundation for generalizations to more flexible structural models that we present subsequently. 

\bigskip


\noindent {\bf{The Bayes Estimator:}} We consider a latent Gaussian vector $\B{\mu}_{n \times 1}= (\mu_1, \ldots, \mu_n)$ where, $\mu_{i} \stackrel{\text{iid}}{\sim} \mathcal{N}(0, A^2)$; that gives rise to an observable Gaussian vector $\B{z}=(z_1, \ldots, z_n)$ such that $z_i|\mu_{i} \sim \mathcal{N}(\mu_{i}, 1)$ for all $i$. 
Note that the posterior distribution of $\B{\mu} | \B{z}$ is given by a $n$-dimensional multivariate Gaussian with mean 
$B^2 \B{z}$ and covariance $B^2\B{I}$, i.e., 
$\B{\mu} | \B{z} \sim \mathcal{N}_n(B^2 \B{z}, B^2\B{I})$  where,  $B^2 = \frac{A^2}{1+A^2}.$
Recall that the Bayes estimator is the mean of the posterior $\B{\mu} | \B{z}$ and given by:
\begin{equation}\label{EB}
\hat{\B{\mu}}^{\text{Bayes}} = \mathbb{E}(\B{\mu} | \B{z})
= \left( 1- {1}/{(1+A^2)}  \right) \B{z}.
\end{equation}
We remind the reader that the Bayes estimator shrinks each observation towards the mean $0$ of the prior distribution -- this is to be contrasted with the usual maximum likelihood (ML) estimator, $\hat{\B\mu}^{\text{ML}} = \B{z}$
that does not shrink $\mu_{i}$'s.
The Bayes estimator has smaller risk than the ML estimator (Theorem~\ref{thm-1}).

\bigskip

\noindent {\bf{Empirical Bayes (EB) estimator:}}
The Bayes estimator depends upon the unknown hyper-parameter $A^2$; which needs to be estimated from data. The
``Empirical Bayes" (EB) framework~\cite{efron2012large} achieves this goal by using a data-driven plug-in estimator for $A^2$ in~\eqref{EB} -- this leads to an EB estimator for $\B\mu$.

The basic EB framework obtains an unbiased estimator for $A^2$ based on the marginal distribution of 
$\B{z} {\sim} \mathcal{N}_{n} \left( \B{0},  \left(1+ A^2\right) \B{I} \right).$
Using standard properties of Gamma and inverse-Gamma distributions, it follows that 
$(n-2)/S$ (where, $S = \sum_{i=1}^n z_i^2$) is an unbiased estimator
of $\tfrac{1}{A^2+1}$. This leads to an EB estimate $\hat{\B{\mu}}^{\text{EB}} = \left( 1- \frac{n-2}{S}  \right) \B{z}$,
which has smaller risk (Theorem~\ref{thm-1}) than the ML estimator.

\begin{theorem}\label{thm-1}\cite{efron2012large}
	For $n\ge 3$, the EB estimator $\hat{\B\mu}^{ \text{EB} }$ has smaller risk (defined as $R(\hat{\B\mu}) := \mathbb{E} ( \| \B\mu - \hat{\B\mu} \|_{2}^2)$) than the ML estimator $\hat{\B{\mu}}^{\text{ML}}$, i.e., $R ( \hat{\mu}^{ \text{EB} }) < R ( \hat{\B\mu}^{\text{ML}} )$ for any $\B\mu$.
The risks of the EB and Bayes estimators are comparable, with a relative ratio:
    $\tfrac{R(\hat{\B\mu}^{ \text{EB} })-R( \hat{\B\mu}^{ \text{Bayes} })}{R( \hat{\B\mu}^{ \text{Bayes} })} = \tfrac{2}{nA^2}.$ 
\end{theorem}
We make the following remarks: (i) Theorem~\ref{thm-1} states that the price to pay for not knowing $A$ is rather small, and as $n$ becomes large the Bayes and EB estimators are similar. 
(ii) Instead of taking an unbiased estimator for $1/(A^2+1)$ as above, one can also take a consistent estimator which might be easier to obtain for more general models (see Section~\ref{sec: model-EB}). For more general models, a good plug-in estimate for $A$ may be obtained based on validation tuning. The framework above provides important guidance regarding a range of good choices of $A$ thereby reducing the computational cost associated with the search for tuning parameters.  

\bigskip

\noindent {\bf{MAP estimation:}}
As an alternative to the Bayes/EB estimator, we can also consider the MAP estimator, a mode
of the posterior likelihood $\B{\mu}|\B{z}$.
Here the MAP estimate ($\hat{\B{\mu}}^{\text{MAP}}$) coincides with the Bayes estimator.
The MAP and Bayes estimators are not the same in general. 
For flexible priors/models, computing Bayes estimators can become computationally challenging and one may need to resort to (intractable) high dimensional MCMC computations.
In these situations (see Section~\ref{sec:EB-cov}), the MAP estimator may be easier to compute from a practical viewpoint. For all models used in this paper, we observe that MAP computation can be tractably performed via convex optimization.


\subsection{Model 1: Modeling user effects}\label{sec: model-EB}
We present a hierarchical shrinkage framework for predicting user-specific session lengths, generalizing the framework in Section~\ref{sec:bayesian}.

Suppose that the log-session lengths of the $i$th user are normally distributed with mean $\mu_i$; and these latent variables $\{\mu_{i}\}_{1}^{N}$ 
are generated from a centered Gaussian distribution.
The (random) $\mu_i$ is the $i$th user effect. This leads to the following hierarchical model:
\begin{equation}\label{sl-framework}
y_{ij} | \mu_i \stackrel{\text{iid}}{\sim} \mathcal{N}\left( \mu_i, \sigma_1^2 \right), i \in [N], \ j  \in [n_i];~~
\mu_{i} \stackrel{\text{iid}}{\sim} \mathcal{N}(0, \sigma_0^2)
\end{equation}
generalizing the model in Section~\ref{sec:bayesian} to the case with multiple replications per user.
The posterior distribution is given by:
\begin{equation*}
\mu_i | \B{y}_i \sim \mathcal{N} \left(\frac{\bar{{y}}_i}{1+ \lambda/n_i}, \frac{\sigma_1^2}{\lambda + n_i} \right)~\forall i, 
\end{equation*}
where, $\lambda = {\sigma_1^2}/{ \sigma_0^2}$ and $\bar{y}_{i}= \sum_{j=1}^{n_i} y_{ij}/n_i$ is the mean of the vector $\B{y}_i$.
The Bayes estimator of $\B\mu$ is given by the posterior mean
$\hat{\mu}_i^{\text{Bayes}} = \frac{\bar{{y}}_i}{1+ \lambda/n_i}$.
Here, the MAP estimator of $\B\mu$ coincides with the 
Bayes estimator as well.
We note that the Bayes/MAP estimators in this example bear similarities with the model in Section~\ref{sec:bayesian} -- we shrink the observed mean of each user towards the global mean of the prior distribution: this lowers the variance of the estimator at the cost of (marginally) increasing the bias. The amount of shrinkage depends upon the number of sessions of the $i$th user via the factor $1+\lambda/n_i$. In particular, the shrinkage effect will be larger for users with a small number of sessions.

\bigskip

\noindent {\bf{Estimating the hyper-parameters:}} The estimators above depend upon hyper-parameters $\sigma_{0},\sigma_1$ via 
$\lambda = \sigma_1^2/\sigma_0^2$, which is unknown and needs to be estimated from data. In the spirit of an EB estimator 
we obtain a plug-in estimator for $\lambda$. 
To this end we use the marginal distribution of $\B{y}_{i}$, which follows
$\mathcal{N}_{n_i}\left( \B{0}, \B{\Sigma}_{n_i} \right)$
where, $\B{\Sigma}_{n_i}$ has
diagonal entries equal to $\sigma_0^2 + \sigma_1^2$ and off-diagonal entries equal to $\sigma_0^2$. Consequently, $\B{y}_i \B{y}_i^T$ is an unbiased estimator for the covariance matrix $\B{\Sigma}_{n_i} $. In particular, if 
$T_i = \| \B{y}_i \|_2^2$ then the estimators
\begin{equation}\label{user-estimator}
\hat{\sigma}_0^2(i) = \frac{( n_{i}\bar{\B{y}}_i)^2 - T_i}{n_i(n_i-1)} \ \ \text{ and } \ \ \hat{\sigma}_0^2(i) + \hat{\sigma}_1^2(i) = \frac{T_i}{n_i} 
\end{equation}
are unbiased estimators of $\sigma_0^2$ and  $\sigma_0^2 + \sigma_1^2$ (respectively). 
To see this, note that $\hat{\sigma}_0^2(i)$ is obtained by taking the average of all the $n_i(n_i-1)$ off-diagonal entries of the matrix $\B{y}_i \B{y}_i^T$. Similarly, $\hat{\sigma}_0^2(i) + \hat{\sigma}_1^2(i)$ corresponds to the average of the diagonal entries.
Estimators in~\eqref{user-estimator} are based solely on 
observations from the $i$th user; and can have high variance if $n_i$ is small (which is the case for less heavy users).
Hence, we aggregate the estimators across all $N$ users to obtain improved estimators of $\sigma_0^2,\sigma_{1}^2$ given by:
{{\begin{equation}\label{var-EB1}
\begin{aligned}
\hat{\sigma}_0^2 =  \tfrac{1}{N} \sum_{i=1}^N \hat{\sigma}_0^2(i)  =  \tfrac{1}{N}\sum_{i=1}^N  \tfrac{( \B{1}_{n_i}^T \B{y}_i)^2 - T_i }{n_i(n_i-1)} \\
\hat{\sigma}_1^2 =  \tfrac{1}{N} \sum_{i=1}^N \hat{\sigma}_1^2(i) = \tfrac{1}{N} \sum_{i=1}^N \tfrac{T_i}{n_i} - \hat{\sigma}_0^2.
\end{aligned}
\end{equation}}}
Using laws of large numbers, one can verify that $\hat{\sigma}_0^2$ (and $\hat{\sigma}_1^2$) are consistent estimators for ${\sigma}_0^2$ (and ${\sigma}_1^2$).  Interestingly, this holds under an asymptotic regime where, $N\rightarrow \infty$ but
$\min_i n_{i}$ remains bounded -- this regime is relevant for our problem since there are many users with few/moderate number of sessions.
We emphasize that even if $N$ is large but the $n_{i}$'s are small, shrinkage plays an important role and leads to estimators with smaller risk than the usual maximum likelihood estimator $\mu^{\text{ML}}_{i}=\bar{\B{y}}_i$ for $i \in [N]$. 
The plug-in estimators suggested above lead to consistent estimators for the Bayes and EB estimators. 
This framework provides guidance regarding the choice of the tuning parameters in practice (and reduces the search-space associated with hyperparameter tuning).



\subsection{Model 2: Modeling with covariates}\label{sec:EB-cov}
We describe a generalization of Model~1 that incorporates user and device-specific covariates (See Section~\ref{sec:datasets} for details). Our hierarchical model is now given by:
\begin{equation}\label{model-2}
\begin{aligned}
y_{ij} | \B{\beta}, \mu_i \stackrel{\text{iid}}{\sim}  \mathcal{N}\left(\B{x}_{ij}^T\B{\beta} + \mu_i, \sigma_1^2\right),  i \in [N], j\in [n_i]\\
\text{where, }~\B{\beta} \sim \mathcal{N}_{d}(\B{0}, \sigma_2^2 \B{I} ),~~~ \mu_i \stackrel{\text{iid}}{\sim} \mathcal{N}(0, \sigma_0^2), i \in [N].
\end{aligned}
\end{equation}
The above represents a generative model with latent variables $(\B\mu,\B\beta)$ -- where, $\B\beta \in \mathbb{R}^d$ denotes a vector of regression coefficients corresponding to the covariates $\B{X}$;
and $\mu_{i}$ explains the residual user-specific effect of user $i$. 
Both the latent variables are normally distributed with mean zero; and
given these parameters, $y_{ij}$'s are normally distributed with mean 
$\B{x}_{ij}^T\B{\beta} + \mu_i$. This model is more structured and also more flexible than Model~1 in that the $i$th user effect has two components: a global regression-based response $\B{x}_{ij}^T\B{\beta}$ (this depends upon both the user and the session); and a residual component $\mu_{i}$. We now derive the EB and MAP estimators.
Let us define $\tilde{\B{\mu}} = \frac{\sigma_2}{\sigma_0} \B{\mu}$
and the latent vector $\B{\gamma}=\left(\B{\beta}, \tilde{\B{\mu}} \right)$. Model~\eqref{model-2} can be reformulated as:
\begin{equation*}
\begin{aligned}
y_{ij} | \B{\gamma}\stackrel{\text{iid}}{\sim} \mathcal{N} \left( \tilde{\B{x}}_{ij}^T\B{\gamma}, \sigma_1^2 \right),\forall i,j;~~\text{ }~~~\B{\gamma} \sim \mathcal{N}_{N+d}\left(\B{0}, \sigma_2^2 \B{I}\right), 
\end{aligned}
\end{equation*}
where, $\tilde{\B{x}}_{ij} \in \mathbb{R}^{d+N}$ is such that its first $d$ entries correspond to $\B{x}_{ij}$, its $(d+i)$th entry is $\sigma_0/\sigma_2$; and all remaining entries are 0.
If $\B{\tilde{X}}_{N_0 \times d+N}$ be the matrix obtained by row concatenation of the $\tilde{\B{x}}_{ij}$'s; then the posterior distribution of $\B{\gamma} | \B{y}$ is given by
$$\B{\gamma} | \B{y} \sim \mathcal{N}_{N+d} \left( \B{H}^{-1}\B{\tilde{X}}^T\B{y} , \sigma^2_2 \B{H}^{-1}\right),$$
where, the matrix $\B{H}= \B{\tilde{X}}^T\B{\tilde{X}} + \alpha \B{I}$  and the regularization parameter $\alpha = \sigma_1^2 / \sigma_2^2$. The Bayes estimate of $\B{\gamma}$ is given as:
\begin{equation*}\label{cov-EB}
\hat{\B{\gamma}}^{\text{Bayes}} = \mathbb{E}(\B{\gamma} | \B{y} ) = \left( \B{\tilde{X}}^T\B{\tilde{X}} + \alpha \B{I}_{d+N}  \right)^{-1}\B{\tilde{X}}^T\B{y} \in \mathbb{R}^{d+N}.
\end{equation*}
$\hat{\B{\beta}}^{\text{Bayes}}$ and $\hat{\B{\mu}}^{\text{Bayes}}$ can be derived from the components of $\hat{\B{\gamma}}^{\text{Bayes}} = \left( \hat{\B{\beta}}^{\text{Bayes}}, \frac{\sigma_2}{\sigma_0} \hat{\B{\mu}}^{\text{Bayes}} \right)$.
In this model, the MAP estimator coincides with the Bayes estimator, and
can be computed as $(\hat{\B{\beta}}^{\text{MAP}}, \hat{\B{\mu}}^{\text{MAP}}) \in \argmin \mathcal{L}_2(\B{\beta}, \B{\mu})$, where,  
$\mathcal{L}_2(\B{\beta}, \B{\mu})$ is the convex function:
\begin{equation}\label{likelihood-l2}
\mathcal{L}_2(\B{\beta}, \B{\mu}):= \sum_{i=1}^N \left\{ \sum_{j=1}^{n_i}  (y_{ij} -  \B{x_{ij}}^T \B{\beta}  - \mu_i )^2  + \lambda \mu_i^2 \right\} + \alpha \|  \B{\beta}  \|_2^2 ,
\end{equation}
and $\lambda={\sigma_{1}^2}/{\sigma^2_0}$,
$\alpha={\sigma_{1}^2}/{\sigma^2_2}$ 
are hyper-parameters. In Section \ref{sec:algo}, we propose Algorithm~1 to minimize Problem~\eqref{likelihood-l2}. 

An empirical Bayes estimator of $(\B{\beta}, \B{\mu})$
can be computed by using data-driven estimators for 
the hyper-parameters. 
We can obtain consistent estimators of the hyper-parameters
following the derivation 
in Section~\ref{sec: model-EB}. Since this derivation \footnote{Note that for Model~\eqref{model-2} the marginal distribution of $\B{y}$ is a multivariate Gaussian with mean zero and 
covariance matrix $\B\Sigma$, which is a function of 
$\{\sigma_{i}\}_0^{2}$ and $\B{X}\B{X}^T$. Following Section~\ref{sec: model-EB}, we have $\mathbb{E}(\B{y}\B{y}^T) = \B\Sigma$. We can then derive consistent estimators of $\{\sigma_{i}\}_0^{2}$ based on functionals of $\B{y}\B{y}^T$ and entries of $\B{X}\B{X}^T$.} is quite tedious, we do not report it here. In practice, we recommend tuning $(\lambda,\alpha)$ on a validation set (where, $\lambda$ is taken to be in the neighborhood of the values suggested by Section~\ref{sec: model-EB} pertaining to Model~1). As we discuss in Section~\ref{sec:compute-scale}, this does not add significantly to the overall computational cost, as our algorithm effectively uses warm-start continuation~\cite{friedman2001elements} across different tuning parameter choices.

We now move beyond the Gaussian prior setup considered so far and consider a Laplace prior on $\B\beta$. In this case, and the models we consider subsequently, Bayes estimators are difficult to 
compute due to high-dimensional integration that require MCMC computations. With computational tractability in mind, we will resort to MAP estimation for these models. 

\bigskip

\noindent {\bf{Laplace prior on $\B{\beta}$:}}
Motivated by $\ell_{1}$ regularization techniques~\cite{tibshirani1996regression} popularly used in sparse modeling, 
we propose a Laplace prior on $\B\beta$ -- the corresponding MAP estimators 
lead to sparse, interpretable 
models~\cite{friedman2001elements,tibshirani1996regression}. 
Here, computing the Bayes estimator becomes challenging and requires MCMC computation.
However, the MAP estimator is particularly appealing from a statistical and computational viewpoint; and 
given by $\left( \hat{\B{\beta}}^{\text{MAP}}, \hat{\B{\mu}}^{\text{MAP}} \right) \in \argmin \mathcal{L}_1( \B{\beta}, \B{\mu})$, where, 
\begin{equation}\label{likelihood-l1}
\mathcal{L}_1( \B{\beta}, \B{\mu}) := \sum_{i=1}^N \left\{ \sum_{j=1}^{n_i}  (y_{ij} -  \B{x_{ij}}^T \B{\beta}  - \mu_i )^2  + \lambda \mu_i^2 \right\}  + \alpha \|  \B{\beta}  \|_1
\end{equation}
is a convex function with hyper-parameters $\lambda, \alpha$. The tuning parameters are chosen based on a validation set. 
Section~\ref{sec:algo} presents an algorithmic framework based on first order convex optimization methods~\cite{nesterov2013gradient,wright2015coordinate} for optimizing Problem~\eqref{likelihood-l1} -- our proposed algorithm leads to significant computational gains compared to off-the-shelf implementations. 

\subsubsection{Nonparametric modeling with covariates}\label{sec:parametric}
The framework presented above is quite modular--it allows for flexible generalizations, allowing a practitioner to experiment with several modeling ramifications, and understand their incremental value (prediction accuracy vis-a-vis computation time) in the context of the particular application/dataset. 

Recall that the basic model put forth by Model~2 is 
$ y_{ij} | \theta_{ij} \stackrel{\text{iid}}{\sim}  \mathcal{N}\left(\theta_{ij}, \sigma_1^2\right)$  where,  $\theta_{ij}  = \B{x}_{ij}^T \B{\beta} + \mu_i$.
 We propose to generalize this linear `link' by incorporating flexible nonparametric models for the covariates, as follows:
\begin{equation}\label{gen-link}
y_{ij} | \theta_{ij} \stackrel{\text{iid}}{\sim}  \mathcal{N}\left(\theta_{ij}, \sigma_1^2\right), \text{ and } 
\theta_{ij}  = f(\B{x}_{ij}) + \mu_i,
\end{equation}
where, $f(\cdot)$ is a flexible nonparametric function of the covariates.
For example, we can train $f$ 
via Gradient Boosting Trees (GBT)~\cite{friedman2001elements} as our non-parametric model\footnote{We also experimented with classical Classification and Regression trees as well as random forests, but the best predictive models were obtained via GBT.}. Trees introduce nonlinearity and higher order interactions among features and can fit complex models. By adjusting tuning parameters like learning rate, maximal tree-depth, number of boosting iterations, etc, they control the bias-variance trade-off and hence the generalization ability of a model.
Given a continuous response $\B{z}_{n \times 1}$ and covariates $\B{U}_{n \times d}$; GBT creates an additive function of the form $f(\B{U}) = \sum_{k} \eta h_{k}(U)$ where, $h_{k}(\cdot)$'s are trees 
of a certain depth and $\eta$ is the learning rate -- the components $\{h_{k}\}$ are learned incrementally via steepest descent on the least squares loss 
$\| \B{z} - f(\B{U}) \|_2^2$
with possible early stopping. This imparts regularization and improves prediction accuracy. 

\bigskip

\noindent{\bf{Summarizing the general framework:}} In summary,  our framework assumes that we have access to an oracle that solves the following optimization problem
\begin{equation}\label{solve-f}
\hat{f} \in \argmin \nolimits_{f} \left\{ \| \B{z} - f(\B{U}) \|_2^2 + \Omega(f) \right\},
\end{equation}
with a regularizer $\Omega(\cdot)$ that restricts the family $f$. Problem~\eqref{solve-f} encompasses the different models that we have discussed thus far: e.g., model~\eqref{likelihood-l1} (here, 
$f(U) = U\B\beta$ and $\Omega(\B\beta) = \alpha\| \B\beta \|_{1}$), model~\eqref{likelihood-l2} (here, 
$f(U) = U\B\beta$ and $\Omega(\B\beta) = \alpha\| \B\beta \|^2_{2}$); and GBT.


For flexible nonparametric models, a MAP estimator can be obtained by 
minimizing the negative log-likelihood of the posterior distribution jointly w.r.t $\B\mu$ and ${f}$.
This entails minimizing the negative log-likelihood of the posterior distribution -- this is given by the function $\mathcal{L}( f, \B{\mu})$ (up to constants) as follows:
\begin{equation}\label{likelihood-cov-ter}
\mathcal{L}( f, \B{\mu}) := 
\sum_{i=1}^N \left\{ \sum_{j=1}^{n_i}  (y_{ij} -  f \left(\B{x}_{ij}\right)  - \mu_i )^2  + \lambda \mu_i^2  \right\} + \Omega(f),
\end{equation}
where, $\lambda = \sigma_1^2 /\sigma_0^2$. 
Section~\ref{sec:algo} presents an algorithmic framework 
for minimizing~\eqref{likelihood-cov-ter} to obtain estimates $\hat{f},\hat{\B\mu}$.  

We note that for the class of the models in Section~\ref{sec:parametric}, it is not clear how to tractably construct and compute Bayes/EB estimators--we thus focus on MAP estimation; and note that the associated tasks can be cast as tractable convex optimization problems. 

\subsubsection{Some Special Cases}\label{special-cases} As we have noted before, an important contribution of this paper is to propose a general modeling framework -- so that a practitioner can glean insights from data analyzing the incremental gains available from different modeling components. To this end, we note that if all the residual user effects are set to zero (i.e, $\mu_i=0$ for all $i$), then we can use covariates alone to model the user effects. In the case of model~\eqref{likelihood-l2} with $\B\mu = \B{0}$ this is referred to as {\bf{\Ridge}}~in Section~\ref{sec:comparisons}.
Furthermore, if we learn $f(\cdot)$ via boosting (with $\B\mu=\B{0}$) then we recover the model proposed in~\cite{SIGIR2017} (denoted as~\BaselineSIGIR~in Section~\ref{sec:experiments}) as a special case. Predictive performances of these models are presented in Section~\ref{sec:experiments}.


\subsubsection{Robustifying against outliers}\label{sec:huber-loss} 
The models described above assume a normal distribution (see~\eqref{gen-link}) --- in reality, to be more resistant to outliers in the data it is useful to relax this assumption to account for heavier tails~\cite{huber2011robust} in the error distribution. To this end, with computational tractability in mind, we propose a scheme which is a simple and elegant modification to our framework, by assuming a stylized decomposition of the link
$\theta_{ij} = f(\B{x}_{ij}) + \mu_{i}$ in~\eqref{gen-link}. To this end, we write
\begin{equation}\label{decomp-1-h}
\theta_{ij} = f(\B{x}_{ij}) + \mu_{i} + s_{ij}
\end{equation}
and place an additional prior on $s_{ij}$'s -- they are all drawn from $\pi_{\delta},$ where, $\pi_{\delta}(u) = \delta\exp(-2\delta |u|)$ is the Laplace density. The MAP estimator for this joint model requires minimizing the following convex function
{ {\begin{equation}\label{likelihood-cov-ter-huber-loss}
\begin{aligned}
\mathcal{L}( f, \B{\mu}, \B{s}) :=& 
\sum_{i=1}^N \left\{ \sum_{j=1}^{n_i}  (y_{ij} -  f \left(\B{x}_{ij}\right)  - \mu_i - s_{ij} )^2 + \lambda \mu_i^2  \right\}  \\
& + \Omega(f) + 2\delta \sum\nolimits_{ij} |s_{ij}|.
\end{aligned}
\end{equation}}}
w.r.t. the variables $f, \B\mu, \B{s}$.
It is not immediately clear why an estimator available from Problem~\eqref{likelihood-cov-ter-huber-loss} has robustness properties. To this end, Theorem~\ref{thm-huber-1} establishes a crisp characterization of Problem~\eqref{likelihood-cov-ter-huber-loss} in terms of minimizing a Huber loss~\cite{huber2011robust} on the residuals
$y_{ij} -  f \left(\B{x}_{ij}\right) - \mu_{i},$
where, the Huber-loss is given by 
$$H_{\delta}(a) = \begin{cases} a^2 & \text{if}~~~~|a| \leq \delta \\
\delta (2 |a| - \delta) & \text{otherwise.}\end{cases} $$
The Huber loss is quadratic for small values of $a$ (controlled by the parameter $\delta$) and linear for larger values -- thereby making it more resistant to outliers in the $y$ space. The Huber loss remains relatively agnostic to the size of the residuals, therefore offering a robust approach to regression~\cite{huber2011robust}. If $\delta$ is small, $H_{\delta}(a)$ resembles the least absolute deviation loss function---this makes it more suitable for the \MAE~metric used for evaluation in Section~\ref{sec:experiments}.
\begin{theorem}\label{thm-huber-1}
Minimizing Problem~\eqref{likelihood-cov-ter-huber-loss} w.r.t $(f, \B\mu, \B s)$ is equivalent to minimizing $\mathcal{L}_{\delta}( f, \B{\mu})$ (below) w.r.t. $(f, \B\mu)$:
\begin{equation}\label{likelihood-cov-ter-huber-loss-1}
\mathcal{L}_{\delta}( f, \B{\mu} ) := 
\sum_{i=1}^N \left\{ \sum_{j=1}^{n_i}  H_{\delta}(y_{ij} -  f \left(\B{x}_{ij}\right)  - \mu_i) + \lambda \mu_i^2  \right\} + \Omega(f)
\end{equation}
\begin{proof}
We use a variational representation of the Huber loss
\begin{equation}\label{huber-1}
{H}_{\delta}(a) = \min_{s \in \mathbb{R}} \psi( s, a):= \left\{ (s-a)^2 + 2\delta|s|  \right\}.
\end{equation}
To derive identity~\eqref{huber-1}, we compute $\hat{s}$ a minimizer of $s \mapsto \psi( s, a)$ in~\eqref{huber-1} via soft-thresholding: $\hat{s} = \sign(a) ( | a| - \delta)_+$ (where, $(\cdot)_+ := \max \{ \cdot, 0\}$). We plug-in the value of $\hat{s}$ into
$\psi(s,a)$ and upon some simplification obtain~\eqref{huber-1}.
The proof of the theorem follows by applying~\eqref{huber-1} to Problem~\eqref{likelihood-cov-ter-huber-loss}, where, we minimize 
$\mathcal{L}( f, \B{\mu}, \B{s})$ wrt $\B{s}$ to obtain criterion~\eqref{likelihood-cov-ter-huber-loss-1} involving $f, \B\mu$ (and not $\B{s}$).
\end{proof}
\end{theorem}
The above development: decomposition~\eqref{decomp-1-h} and hence criterion~\eqref{likelihood-cov-ter-huber-loss} nicely falls within the general hierarchical framework discussed in this paper. In fact all models described before this section can be cast as special instances of Problem~\eqref{likelihood-cov-ter-huber-loss} by setting $\delta = \infty$ and stylized choices of $f(\cdot), \Omega(\cdot)$. Our numerical experiments suggest that a finite nonzero choice of $\delta$ leads to the best out-of-sample prediction performance, thereby suggesting the importance of doing robust modeling in this application. In addition, our model is nicely amenable to the computational methods discussed in Section~\ref{sec:algo}. This further underlines the flexibility of our overall framework -- even if our basic assumption relies on Gaussian errors at the core, simple hierarchical modeling decompositions of the latent variables make it flexible enough to accommodate adversarial corruptions in the data. 


\subsection{Computation via Convex Optimization}\label{sec:algo}
All the estimation problems alluded to above can be cast as convex optimization problems; for which we resort to modern computational methods~\cite{nesterov2013gradient,wright2015coordinate}. 
To compute the estimators mentioned in Section~\ref{sec:EB-cov}, we need to minimize Problem~\eqref{likelihood-cov-ter-huber-loss}. To this end, we use a block-coordinate descent scheme~\cite{wright2015coordinate}: at iteration $t$, we minimize~\eqref{likelihood-cov-ter-huber-loss} w.r.t. $f$, followed by a minimization w.r.t the latent vectors $\B{\mu}, \B{s}$. The algorithm is summarized below.



\bigskip
\noindent \textbf{Algorithm~1: \underline{Block-Coordinate-Descent for MAP estimation}}  \newline
\textbf{Input:} $\B{X},\B{y}$, tuning parameters,  tolerance $\epsilon$; initialization 
${f}^0, \B\mu^0, \B{s}^0$. \newline
\textbf{Output: } An estimate 
$(\hat{f},\hat{\B{\mu}}, \hat{\B{s}})$, minimizing Problem~\eqref{likelihood-cov-ter}

\smallskip
\noindent~({\bf{1}})    Repeat Steps 2 to 5 until  $| L_t - L_{t-1}| / L_{t-1} > \epsilon$ for $t \geq 1$.
	
\noindent~({\bf{2}}) 
Let $\hat{f}^{(t)} \in \argmin_{ f }  \mathcal{L}\left( f, \hat{\B{\mu}}^{(t-1)}, \hat{\B{s}}^{(t-1)} \right)$ be a solution of the optimization problem~\eqref{likelihood-cov-ter} with $\B\mu, \B{s}$ held fixed at $\hat{\B{\mu}}^{(t-1)}, \hat{\B{s}}^{(t-1)}$ respectively -- this is equivalent to solving~\eqref{solve-f} with
    $\B{z} = \B{z}^{(t)}$ where 
    $z_{ij}^{(t)} = y_{ij} -  \hat{\mu}_i^{(t-1)} - \hat{s}_{ij}^{(t-1)} \ \forall i,j$.
	
\noindent~({\bf{3}}) Update the residuals $r_{ij}^{(t)} = y_{ij} - \hat{f}^{(t)}(\B{x}_{ij}), \ \forall i,j.$
Estimate user-specific effects via: $\hat{\B{\mu}}^{(t)} \in \argmin_{ \B{\mu} }  \mathcal{L}\left( \hat{f}^{(t)}, \B{\mu}, \hat{\B{s}}^{(t-1)} \right) $ (with $f,\B{s}$ respectively set to $\hat{f}^{(t)}, \hat{\B{s}}^{(t-1)}$). This is a closed-form update: 
$\hat{\mu}_i^{(t)} = \tfrac{1}{n_i+ \lambda} \sum_{j=1}^{n_i} \left( r_{ij}^{(t)} - s_{ij}^{(t-1)}\right), \ \forall i$.

\noindent~({\bf{4}}) Update the vector: $\hat{\B{s}}^{(t)} \in \argmin_{ \B{s} }  \mathcal{L}\left( \hat{f}^{(t)}, \hat{\B{\mu}}^{(t)}, \B{s} \right)$, corresponding to the sparse corruptions. The closed form update is $\hat{s}_{ij}^{(t)} = \sign(\eta^{(t)}_{ij})(|\eta^{(t)}_{ij}| - \delta)_+$;
where, $\eta^{(t)}_{ij}=r_{ij}^{(t)} - \mu_i^{(t)}\ \forall i,j$.
   
\noindent~({\bf{5}}) Set the value of $L_{t+1} = \mathcal{L} \left(\hat{f}^{(t)}, \hat{\B{\mu}}^{(t)}, 
\hat{\B{s}}^{(t)}\right)$.

\bigskip

Algorithm~1 applies to a fixed choice of the hyper-parameters. 
We need to consider a sequence of hyper-parameters to obtain the best model based on the minimization of prediction error (see Section~\ref{sec:experiments}) on a validation set.
In the case of models~\eqref{likelihood-l2} and~\eqref{likelihood-l1} 
estimates of $\B\beta$ can be computed over a grid of parameters 
by using warm-starts across different tuning parameters--to this end, the EB estimators provide a good ballpark estimate of relevant tuning parameters. 
Section~\ref{sec:compute-scale}
describes specialized algorithms that are found to speed up the computations pertaining to models~\eqref{likelihood-l2} and~\eqref{likelihood-l1} when compared to off-the-shelf implementations of these algorithms.
We note that GBT does not benefit from warm-start continuation across hyper-parameters. 
For Algorithm~1, we use a stopping criterion of $\epsilon=0.01$ (Step~1) in the experiments.

\subsection{Computational Considerations}\label{sec:compute-scale}
We consider certain algorithmic enhancements for Algorithm~1 that lead to important savings when the number of sessions become large (of the order of millions).  
We focus on the critical Step~2 of Algorithm~1 when $f(U) = U\B\beta$ and $\Omega(\B\beta)$ corresponds to the ridge or $\ell_{1}$ regularization -- this leads to
a problem of the form:
\begin{equation}\label{lasso-ridge}
\min\nolimits_{\B\beta} \left\{ \| \B{z}^{(t)} - \B{X} \B{\beta} \|_2^2 +  \Omega(\B{\beta} ) \right\}.
\end{equation}
where $\Omega(\B{\beta}) \in \left\{ \alpha \| \B{\beta} \|_2^2, \ \alpha \| \B{\beta} \|_1   \right\}$. Indeed, in these instances, we found out that the default implementation of {Python}'s \texttt{scikit-learn} package~\cite{scikit-learn} was prohibitively slow for our purpose, and hence careful attention to 
algorithmic details seemed necessary (details below).
We derived new algorithms for~\eqref{lasso-ridge} with an eye towards caching numerical linear algebraic factorizations, exploiting warm-starts, etc; as we describe below.
\subsubsection{$\ell_2$ regression subproblem}\label{sec:ridge} When $\Omega( \B{\beta})= \alpha \| \B{\beta} \|_2^2$,  the ridge estimator has an analytical expression:
\begin{equation}\label{ridge}
\hat{\B{\beta}}^{\text{R}} = \left( \B{X}^T\B{X} + \alpha \B{I}_{d}  \right)^{-1}\B{X}^T\B{z}^{(t)},
\end{equation}
which needs to be computed for several tuning parameters, and iterations. To reduce the computational cost, we obtain an equivalent expression for $\hat{\B{\beta}}^{\text{R}}$ 
via the eigendecomposion of $\B{X}^T\B{X} \in \mathbb{R}^{d \times d}$ given by, 
$\B{X}^T\B{X} = \B{V} \B\Gamma \B{V}^T$, with $\B\Gamma$ being a diagonal matrix with eigenvalues $\{\gamma_{i}\}_{1}^{d}$ -- this has a cost of $O(d^3)$ in addition to the $O(N_0 d^2)$ cost of computing $\B{X}^T\B{X}$ (and they can both be done once, off-line).
This leads to 
$\hat{\B{\beta}}^{\text{R}}= \B{V} \B\Gamma \B{X}^T \B{z}^{(t)}$ which can be computed with cost $O(N_0 d + d^2)$. In our experiments (Section~\ref{sec:experiments}) $d$ is small, which leads to a cost that is linear in $N_0$. The predicted values $\B{X}\hat{\B\beta}^{\text{R}}$ can be computed with an additional cost of $O(N_0 d)$.
Note that computing estimator~\eqref{ridge} for different values of the tuning parameter $\alpha$ does not require additional eigendecompositions -- this is critical in making the overall algorithm efficient, especially when training across multiple values of the hyper-parameter.

\subsubsection{$\ell_1$ regression subproblem}\label{sec:lasso}  When $\Omega(\B{\beta})= \alpha \| \B{\beta} \|_1$, Problem~\eqref{lasso-ridge} becomes equivalent to a Lasso estimator -- we emphasize that \texttt{scikit}-\texttt{learn}'s implementation of 
Lasso became rather expensive for our purposes since it could not effectively exploit warm-starts and cached matrix computations. This motivated us to consider our own implementation, based on proximal gradient descent~\cite{FISTA,nesterov2013gradient}. To this end, 
since $N_0 \gg d$, we precomputed\footnote{Note that this computation is also required for the ridge regression model.} $Q:=\B{X}^T\B{X}$ and considered a $d$-dimensional quadratic optimization problem of the form:
\begin{equation}\label{l1-qp}
\min\nolimits_{\B\beta}~~F(\B\beta):=\B\beta^T Q \B\beta - 2\langle \B{X}^{T} \B{z}^{t}, \B\beta \rangle  + \alpha \| \B\beta \|_{1}, 
\end{equation}
where, the smooth part of $F(\B\beta)$ has $C$-Lipschitz-continuous gradient -- that is, it satisfies $\| \nabla F(\pmb{\gamma}) - \nabla F(\B{\beta}) \|_2 \le C \|\B{\gamma} - \B\beta\|_2, \forall\B{\gamma},\B{\beta}$ for $C=2\max_{i} \gamma_{i}$ (recall, $\gamma_{i}$'s are eigenvalues of $Q$). A proximal gradient algorithm for~\eqref{l1-qp} performs the following updates:
\begin{align} \label{key-algortihm1}
\begin{split}
{\B{\beta}_{k+1}} 
& \in \argmin\nolimits_{ \B{\beta}  }\left\{ \frac{1}{2} \left\| \B{\beta} - \left(\B{\beta}_{k} - \frac{1}{L} \nabla F(\B{\beta}_{k}) \right) \right\|_2^2 + \frac{\alpha}{L} \|  \B{\beta}  \|_1 \right\}
\end{split}
\end{align}
till convergence. Note that ${\B{\beta}_{k+1}}$ can be computed via soft-thresholding, i.e., 
${\B{\beta}_{k+1}} = \mathcal{S}_{\alpha/L} (\B\beta_{k} - \frac{1}{L} \nabla F(\B{\beta}_{k}) )$ where, for a vector $\B{a} \in \mathbb{R}^d$ the $i$th coordinate of 
the soft-thresholding operator $\mathcal{S}_{\tau} (\B{a} )$ is given by 
 $\text{sgn}(a_i) \max \{ |{a}_i| - \tau, 0 \}$. 
 Note that the objective function $F(\B\beta)$ is 
 strongly convex\footnote{Note that $F(\B\beta) - \rho/2\| \B\beta \|_{2}^2$ is convex for $\rho= 2\min_{i} \gamma_{i}$, i.e., 
 the minimum eigenvalue of $Q$ -- this means that $F(\B\beta)$ is strongly convex with strong convexity parameter $\rho$.}; and hence sequence $\B\beta_{k}$ converges to an $\kappa$-suboptimal solution to Problem~\eqref{l1-qp} in $O(\log(\tfrac{1}{\kappa}))$ iterations~\cite{nesterov2013gradient} -- i.e., it enjoys a linear convergence rate.
Every iteration of~\eqref{key-algortihm1} has a cost of $O(d^2)$ (arising from the computation of $\nabla F(\B\beta)$ and the soft-thresholding operation). 
In addition, computing $\B{X}^T\B{z}^{(t)}$ costs $O(N_0 d)$ (this is computed once at Step~2 of Algorithm~1). 
Problem~\eqref{l1-qp} needs to be computed for several tuning parameters and iterations (of Algorithm~1) -- this does not add much to the overall run-time as the proximal gradient algorithm can be effectively warm-started -- this is found to speed-up convergence in practice. 



\subsubsection{GBT subproblem}\label{GBT-subproblem} When the optimization in Step~2 involves performing GBT, the runtimes increase substantially (See Section~\ref{sec:efficiency}).
Unlike the models in Sections~\ref{sec:ridge},\ref{sec:lasso}; GBT is computationally intensive and needs to be done for every iteration of 
Algorithm~1. Unlike the optimization based algorithms for $\ell_{1}/\ell_2$ regression as described above, GBT does not naturally accommodate warm-starts across iterations, and/or tuning parameters. 

\section{Experiments}
\label{sec:experiments}
%

Here, we evaluate the effectiveness of our prediction model with respect to several baselines and state of the art session length prediction solutions. We proceed by describing our datasets, our evaluation framework, the comparisons, and then present the results.   

\subsection{Datasets}
\label{sec:datasets}
We used two different real world datasets of users listening to music, namely, \pandora~and \lastfm. \pandora~is a sample of user interaction data from a major music streaming service
in United States, 
and \lastfm~is a publicly available dataset from last.fm \cite{OscarBook2010}. We defined the user sessions as periods of continuous listening, interrupted if the user stop or pause the music for more than $30$ minutes \cite{SIGIR2017}. For \pandora~we gathered data from a small subset of \pandora~users for a period of $3$ months (February-May $2016$) resulting in $3,976,561$ sessions \footnote{Due to confidentiality we can not report the number of users for this dataset.}.
\lastfm~public dataset was gathered between $2004$ to $2009$ and it contains $911,770$ sessions for $1,000$ different users. Table \ref{table:pandora-norm-stats} reports some statistics about the user session length in the two datasets. For the log values, we first take the log transform of the raw data, as mentioned in the modeling part, and then normalize. An interesting finding is that mean and median are quite different for the raw data in both datasets. In fact, as reported in \cite{SIGIR2017}, Weibull distributions give a better fit to user session lengths, while after a log-transformation, the data can be reasonably modeled via normal distributions, which is what our modeling framework requires for tractable inference.

\begin{table}[t]
	\centering
	\caption{\small{Summary statistics of normalized user session lengths in the two datasets. The upper half are on the normalized raw session lengths. The bottom half are on the normalized log session lengths.}}
    \smallskip
		\label{table:pandora-norm-stats}
		\begin{tabular}{l|l|l}
            \toprule
            \toprule
			Stats & \pandora & \lastfm \\
			\toprule
			25th quantile(raw) & $0.008$ & $0.009$ \\
			median(raw) & $0.021$ & $0.029$ \\
			mean(raw) & $0.044$ & $0.060$ \\
			75th quantile(raw) & $0.049$ & $0.069$ \\
            \toprule
			25th quantile(log) & $0.57$ & $0.59$\\
            median(log) & $0.66$ & $0.69$\\
            mean(log) & $0.65$ & $0.62$\\
            75th quantile(log) & $0.74$ & $0.76$\\
			\bottomrule
            \bottomrule
		\end{tabular}
\end{table}

\bigskip

\noindent {\bf{Feature Engineering.}} For all the sessions in \pandora~we create two kinds of features, namely, user-based and contextual as in \cite{SIGIR2017}. Table \ref{table:features} reports some of the co-variates used in our models.
As user-based features we consider \textit{"gender (the gender of the user), age (the age of the user), subscription\_status (whether the user is ad-supported)"}, these features are fixed for a given user. As contextual features we consider \textit{"device (the device used for the session), network (the type of network used for the session), absence\_time (time elapsed since the user's previous session), previous\_duration (the duration of the user's previous session)"}.

We refine this set of features to include additional contextual features to \cite{SIGIR2017}, this is mainly to lower the variance of the past sessions, and introduce non-linearity. We consider as additional features  \textit{"avg\_user\_duration( average user session length in training set), log\_avg\_user\_duration (logarithm of avg\_user\_duration), log\_absence\_time (logarithm of absence\_time), log\_previous\_duration (logarithm of previous\_duration), session\_time (whether the user session started in morning or afternoon)"}. For \lastfm~dataset the \textit{"age, subscription\_status, device, network} are missed. 

\begin{table}[t]
	\centering
	\caption{\small{ Example of user-based and contextual features used in the models. }}
    \smallskip
		\label{table:features}
		\begin{tabular}{l|l}
            \toprule
            \toprule
			Feature & Description \\
			\toprule
			 gender &   gender of the user\\
			 age &  age of the user \\
			 subscription\_status &  whether the user is ad-supported\\
             \toprule
			 device &   device used for the session\\
			 network &  type of network used for the session\\
			 absence\_time & time elapsed since the  previous session \\
			 previous\_duration &   duration of the  previous session\\
			 avg\_user\_duration &  average user session length (training)\\
             session\_time & session started in morning or afternoon\\
            \bottomrule
            \bottomrule
		\end{tabular}
\end{table}



\subsection{Evaluation}
We sort our dataset by chronological order, use the first $80\%$ for the training set, $10\%$ for the validation set, and the rest $10\%$ for the test set.
Additionally we require each user in the validation or test set to appear at least once in the training set. 
The final datasets for \pandora~and \lastfm~have respectively in total $3,949,137$, and $713,089$ sessions. For the models that need parameter tuning, we first train the models on the training set for each set of the parameters. Then we use the validation set to pick the best set of parameters. Finally, we use that set of parameters for training on the combined set of training and validation, and predict on the test set.
For the evaluation metric of our session length prediction model, we use \textit{Normalized Mean Absolute Error}  measured in seconds, averaged over all the test sessions and normalized by the \Baseline~model which by our definition has \MAE~$= 1$. \MAE~is a good metric due the possibility of important errors resulting from very large session length. More formally, let $|S_{test}|$ be the number of sessions in the test set and $\tilde{y}_{ij}$ be the time spent by user $i$ on his $j$th session, where $j$ is a test session of user $i$, and $\tilde{y}^{p}_{ij}$ be the predicted value then:
\begin{center}
		$\MAE = \frac{1}{|S_{test}|} \sum_{(i,j) \in S_{test}}{|\tilde{y}^{p}_{ij} - \tilde{y}_{ij}|}$
\end{center}

\subsection{Comparisons}
\label{sec:comparisons}
We compare our model with several baselines and state of the art methods. In particular we have considered the  following: 
%
\begin{description}

\item \Baseline. The baseline model is the per-user mean session length, i.e., we compute for each user the mean session length in the training set and use the value as a prediction value for all the test sessions of the same user.

	\item \XGBoost. This corresponds to a Gradient Boosting Model \cite{xgboost2016} run on basic features to predict session-length. We do not consider the log-transformation.
    \item \BaselineSIGIR. This is method in \cite{SIGIR2017} that is using a modified version of boosting algorithm. Our tuned models have a number of trees in $\left\{10,15,50,100\right\}$, with depth $\left\{6,10\right\}$ and use a learning rate in $\left\{0.1,0.05\right\}$. 
\end{description}
\Baseline~can be interpreted as a natural baseline 
and \BaselineSIGIR~is the state of the art in this particular application. 
Among the models proposed in this paper (cf Section \ref{sec:model}), we consider the following in the experiments:
\begin{description}
    \item \ModelOne. This is the model described in Section \ref{sec: model-EB}, where we don't use any covariates. All the parameters of this model were derived using parameter estimation described in \ref{sec: model-EB}.
    \item \Ridge. This is the Ridge estimator defined in Section~\ref{special-cases}, i.e., we perform a ridge regression only on covariates. We take
    50 values of the tuning parameter (as per Section~\ref{sec:EB-cov}).
    \item \ModelTwoLTwo. This is the Bayes (which is also the MAP) estimator for the model presented in Section~\ref{sec:EB-cov} with an $\ell_2^2$ regularization on $\B{\beta}$. We run Algorithm~1 (Section~\ref{sec:ridge}) on a 2D grid of tuning parameters $(\alpha, \lambda)$ with 500 different values (Section~\ref{sec:EB-cov}). 
    \item \ModelTwoLOne. This is MAP estimator model presented in Section~\ref{sec:EB-cov} with the use of an $\ell_1$ regularization on $\B{\beta}$. We use Algorithm~1 (Section~\ref{sec:lasso}) for computation, and take 500 values of the 2D grid of tuning parameters (Section~\ref{sec:EB-cov}). 
    \item \ModelTwoGBT. This model uses Gradient Boosting Trees (GBT) to compute the MAP estimator~\eqref{likelihood-cov-ter} via Algorithm~1 (Section~\ref{GBT-subproblem}). We use the same sequence of tuning parameters as in \BaselineSIGIR~and a sequence of 10 $\lambda$ values in $[1, 10]$.
    \item \ModelThreeLTwo. This is the extension of \ModelTwoLTwo~
   to criterion~\eqref{likelihood-cov-ter-huber-loss} (or equivalently~\eqref{likelihood-cov-ter-huber-loss-1} with the Huber loss)\footnote{For \ModelThreeLTwo~and \ModelThreeGBT, we take 7 values of $\delta \in [0.1, 10]$.}.
     \item \ModelThreeGBT. This is the extension of \ModelTwoGBT~to criterion~\eqref{likelihood-cov-ter-huber-loss} (or equivalently~\eqref{likelihood-cov-ter-huber-loss-1} with the Huber loss).

\end{description}
Similar to the \Baseline~ model, \ModelOne~does not consider covariates. \ModelOne~however, performs shrinkage on the user-specific effects, and thus any gain in predictive accuracy (as evidenced in Table~\ref{table:results}) is due to shrinkage. 
\Ridge~considers only covariates and does not include the residual user-specific effects. The rest of the models use features regarding the context and user, and additional user-specific effects. All versions of \sf{Model3}~allow us to build models that are less sensitive to outliers, hence any performance boost in \MAE~ over its \sf{Model2}-counterpart can be attributed to robustness. As described in Section \ref{sec:related} we do not have censored data, and we are interested in making point predictions on user session-lengths, therefore survival analysis models are not suitable for our scenario --- hence they are not included in our comparisons. 

\subsection{Effectiveness} 
We report the results regarding the effectiveness of our model.
Table \ref{table:results} reports the results of the Normalized \MAE~on all the models in Section \ref{sec:comparisons}. By  borrowing strength across users, \ModelOne~improves over the \Baseline~even without using any covariate-information.  
\BaselineSIGIR, the model presented in \cite{SIGIR2017}
is benefiting from the usage of the covariates, and it is clearly better than \ModelOne. \XGBoost~which relies (solely on) covariates, has poor performance on both the datasets.
\ModelTwoGBT~by combining hierarchical shrinkage with flexible modeling of covariates, reaches a significantly lower \MAE~than \BaselineSIGIR. This observation shows the importance of the user effect in our hierarchical modeling framework. 
\textsf{Model2-L2} is performing quite well in all the datasets and only considers $2$ hyper-parameters.
We did not observe any gain in MAE by using an $\ell_1$ penalization, though the models were sparse (in $\B\beta$) when compared to $\ell_2$ regularization. \ModelThreeGBT~has the lowest \MAE~for all the datasets --- thereby suggesting the usefulness of using a robust model for training purposes. For the \pandora~dataset, the robustification strategy leads to good improvements: \ModelThreeLTwo~ has \MAE~$0.891$ whereas, its non-robust counterpart \ModelTwoLTwo~ has \MAE~$0.911$. Further improvements are possible by using 
nonparametric modeling of the covariates.
Overall, our \ModelThreeGBT~seems to be the best in terms of prediction in both datasets. \ModelTwoLOne~or \ModelTwoLTwo~are close to \BaselineSIGIR~for \pandora~and better for \lastfm~
and as we see in Table \ref{table:efficiency} they are actually much faster in training time. 




\begin{table}[t]
	\centering
	\caption{\small{Normalized MAE on test set for our model compared to the baselines and state of the art.}}
    \smallskip
		\label{table:results}
		\begin{tabular}{l|l|l} 
        	\toprule
            \toprule
			 Models & \MAE~\pandora & \MAE~\lastfm \\
			\midrule
			\Baseline - no covariates & $1.0$ & $1.0$ \\
    \XGBoost & $1.005$ & $0.862$ \\
    \BaselineSIGIR & $0.910$ & $0.826$ \\
			\ModelOne  - no covariates & $0.936$ & $0.830$\\
			
			\Ridge & $0.921$ & $0.828$\\
			\ModelTwoLOne & $0.911$ & $0.824$\\
			\ModelTwoLTwo & $0.911$ & $0.824$ \\
            \ModelThreeLTwo & $0.891$ & $0.822$ \\
			\ModelTwoGBT & $0.878$ & $0.812$\\
            \ModelThreeGBT & \textbf{0.871} & \textbf{0.811}\\
			\bottomrule
            \bottomrule
		\end{tabular}
\end{table}

\bigskip

\noindent \textbf{Feature Importance.}  By centering and normalizing the columns of the matrix of covariates $\B{X}$, the absolute values of the coefficients of $\hat{\B{\beta}}$ for \Ridge~or \ModelThreeLTwo~suggest the relative importances of the features. Table \ref{table:features} reports the highest absolute values of the coefficients for the \Ridge~estimator and for \ModelThreeLTwo~estimator for \pandora~dataset.
\begin{table}[t]
	\centering
	\caption{{\small{Feature importance for \pandora~dataset considering highest absolute value for \textsf{Ridge} and \textsf{Model3-L2} (we centered and normalized all the features first).} }}
    \smallskip
    \label{table:features}
    \begin{tabular}{l|l|l}
        \toprule
		\toprule
		\Ridge&\textit{log\_avg\_user\_duration}& 0.516\\
		&\textit{absence\_time}& $0.430$ \\
		&\textit{avg\_user\_duration}& $0.064$ \\
		&\textit{device}=smartphone& $0.045$ \\
		&\textit{device}=Web& $0.042$ \\
		\midrule
		\ModelThreeLTwo&\textit{device}=smartphone&$0.097$ \\
		&\textit{device}=Web&  $0.088$ \\
		&\textit{avg\_user\_duration}& $0.085$ \\ 
		&\textit{absence\_time}& $0.069$ \\
         &\textit{log\_avg\_user\_duration}& $0.068$ \\  
		\bottomrule
        \bottomrule
	\end{tabular}
\end{table}
Device and time-related features appear as the most relevant features. The two most important features for \Ridge~correspond to logarithm of the average user session length in training set and the absence time since last session. In addition, considering user effect on \ModelThreeLTwo~lowers the magnitude of time-related features, even though they still appear in the top ones.

\noindent \textbf{Performance Breakdown by Sessions per User.}~~
We perform a break down of users into three different types by quantiles of number of sessions per user in the training set. Table \ref{table:error-user} reports the normalized \MAE~for these three groups. The results show a monotonic decrease in terms of gain in performance (with respect to the \Baseline) for all the models with number of sessions per user. This serves as a validation of an important message presented through our modeling framework -- shrinkage is more critical for less active users when compared to the \Baseline. In fact, even for the more frequent users, we observe that shrinkage helps when compared to the \Baseline~(but the gains are less pronounced).
Both \ModelTwoGBT/\ModelThreeGBT~outperform the state of art \BaselineSIGIR~for all three cutoffs chosen. 
The gains obtained by \ModelThreeGBT~over \ModelTwoGBT~for \pandora~dataset can be primarily attributed to the robustness to outliers.
\ModelTwoGBT/\ModelThreeGBT~perform better than \BaselineSIGIR~by modeling the user-specific effects -- this gain seems to be most prominent for heavier users for both \pandora~and \lastfm~datasets. 

\begin{table}[t]
	\centering
	\caption{\small{Normalized MAE, restricted to people in the first decile, the first two deciles or the last $8$th deciles of the training set. (Hierarchical) shrinkage has a more prominent effect over the \Baseline~model (with MAE=1) for users with fewer sessions.} }
    \label{table:error-user}
		\begin{tabular}{llll}
			\toprule
            \toprule
			Model & $<q_{10}$ &$<q_{20}$ & $>q_{20}$ \\
			\toprule
            \multicolumn{4}{c}{\pandora}\\
            \toprule
			\ModelOne & $0.860$ &  $0.876$  & $0.938$  \\
			\Ridge & $0.790$ & $0.809$ & $0.925$ \\
			\BaselineSIGIR & $0.780$  & $0.806$  & $0.904$  \\
            
            \ModelTwoGBT & $0.778$ &  $0.804$ &  $0.881$ \\
            \ModelThreeGBT & $\textbf{0.767}$ &  $\textbf{0.792}$ &  $\textbf{0.875}$ \\

            \toprule
            \multicolumn{4}{c}{\lastfm}\\
            \toprule
			\ModelOne & $0.610$ &  $0.798$  & $0.834$  \\
			\Ridge & $\textbf{0.606}$ & $0.789$ & $0.833$ \\
			\BaselineSIGIR & $0.622$  & $0.779$ &  $0.829$\\
            \ModelTwoGBT & $\textbf{0.606}$ &  $\textbf{0.775}$ &  $0.817$ \\
            \ModelThreeGBT & $0.609$ &  $\textbf{0.775}$ &  $\textbf{0.816}$ \\
			\bottomrule
            \bottomrule
		\end{tabular}
\end{table}

\subsection{Efficiency} \label{sec:efficiency}
We further investigate the efficiency of our model by looking at the training time and time of prediction for our best models and state of art solution. We implemented everything using \textsc{python} 
\footnote{Our code will be released, we are removing the link to github due to double blind process.}.
We run the experiments in a MacBook Pro with $2.7$GHz Intel Core i$5$ with 8 GB of RAM. 
For each model we fixed a set of parameters to tune. We then run each model sequentially, each time with different fixed parameters. 
Table \ref{table:efficiency} reports the average running time (across the number of runs done for tuning purposes) and the time to predict all the test instances for our best models and \BaselineSIGIR. The most important thing to note here is that \ModelThreeLTwo~can be trained between $20$ to $30$ times faster than any tree based method such as \BaselineSIGIR, and it has an \MAE~that is $1$\% to $2$\% better than \BaselineSIGIR. 
This is mainly because the \ModelThreeLTwo~(and \ModelTwoLTwo) computations benefit from warm-starts (see Section \ref{sec:algo}), so the increase in number of tuning parameters do not increase the run time as in the case of \BaselineSIGIR~and \ModelThreeGBT~that are based on Gradient Boosting Trees. 
They are also at least $30$ to $500$ times faster in prediction time, and therefore they can be used for real-time prediction with time constraints. 
While \ModelThreeGBT~ is slow in training time,
it shows better \MAE~performance compared to other methods. However, we note that these models are meant for off-line training in the context of the application herein--while it is important to have algorithms that are fast, they are not of critical importance as in tasks where real-time learning is of foremost importance. 

\begin{table}[t]
	\centering
	\caption{\small{Average training time, time of prediction for our best models and the state of the art. We report for each model the effectiveness (in terms of \MAE), training time (in seconds), and average prediction time (in seconds), averaged across the tuning parameters. The GBT-based models have longer training and prediction times, when compared to models that are not based on GBT.}}
    \smallskip
        \label{table:efficiency}
		\label{table:efficiency}
		\begin{tabular}{l|c|c|c} 
        	\toprule
            \toprule
			 Models&\MAE&Train. Time & Pred. Time\\
            \midrule
			\multicolumn{4}{c}{\pandora}\\
            \midrule
            \BaselineSIGIR &  $0.910$ & $731$ & $7.70$ \\
			\ModelTwoLTwo& $0.911$ & $17$ & $0.24$ \\
            \ModelThreeLTwo & $0.891$ &  $36$ &   $0.24$ \\
            \ModelThreeGBT & $0.871$ &  $1885$ &  $9.71$ \\
            \midrule
			\multicolumn{4}{c}{\lastfm}\\
            \midrule
            \BaselineSIGIR &  $0.826$ &  $22$ &   $0.53$ \\
            \ModelTwoLTwo & $0.819$ & $1$  & $0.001$ \\
			\ModelThreeLTwo & $0.819$ & $5$ &  $0.001$ \\
            \ModelThreeGBT & $0.812$ &  $66$ &  $0.64$ \\
			\bottomrule
            \bottomrule
		\end{tabular}
\end{table}


\section{Related work}
\label{sec:related}
%
Session length is an important metric serving as a proxy for user engagement. 
Therefore the solutions and evaluation are tailored to similar duration based engagement metric such as dwell-time prediction. 

\noindent \textbf{Dwell-time.} Liu \emph{et al.} in \cite{Weibull2010DwellTime} presented one of the first studies on dwell-time for web search. Kim \emph{et al.} in \cite{Microsoft2014satisfaction} proposed a dwell-time based user satisfaction prediction model in the web search context. Lalmas \emph{et al.} in \cite{Lalmas2015} proposed new way to improve ad ranking. Barbieri  \emph{et al.} in \cite{Barbieri2016}  propose to use survival forest \cite{randomsurvival2008} using landing page and user feature in the ad context to estimate the dwell-time and incorporate it in the ad ranking system as a quality score. 
Vasiloudis \emph{et al.} in \cite{SIGIR2017} presented recently a first session length prediction model in the music streaming context using survival analysis and gradient boosting \cite{xgboost2016}. In that work, the authors show that in the case of media streaming services the probability of a session to end evolves 
differently for different users. In particular $44\%$ of the users exhibit “negative-aging” length distributions, i.e. sessions that become less likely to end as they grow longer. 
Although not directly comparable, we report that this percentage is going to $98.5\%$ for dwell-time on a web page after a search, i.e., after clicking on a search result the more you stay the more you are likely to stay on the clicked page. 
Finally, Jing \emph{et al.} in \cite{Smola2017} presented a neural network based model combined with survival analysis for recommendation purpose and absence time prediction at the same time. 
In our work we compared against the most recent related work done by Vasiloudis \emph{et al.} \cite{SIGIR2017}. It is worth mentioning that while dwell-time can benefit from survival analysis, because a user can click on a search engine result and never turn back, in our case we don't have censored data, therefore our is a regression problem. 



\noindent \textbf{Empirical Bayes and MAP estimation:}
The statistical models proposed in this paper are inspired by Empirical Bayes methods that are 
well-known in statistics community, dating back to \cite{Robbins1955,EfronMorris1972}. In theory, when the model is true, these estimators are known to lead to estimators with better prediction accuracy when compared to (unregularized) maximum likelihood estimators. Empirical Bayes estimators offer an appealing trade-off between frequentist and Bayesian modeling~\cite{efron2012large} --- in that expensive MCMC computations may be avoided by a clever combination of guided hyperparameter tuning and numerical optimization (to obtain MAP estimates for example). We also consider 
more flexible models wherein MAP (maximum a posteriori) estimation becomes a pragmatic choice from a computational viewpoint.
To our knowledge, this is the first time such a  methodology is used in the context of user session length prediction. 
\section{Conclusions and future work}
\label{sec:conclusion}
\balance
%
In this paper, we presented a new hierarchical modeling framework, inspired by core Bayesian modeling principles to predict the amount of time people will spent in a streaming service, and in particular listening to streaming music. We also propose modern convex optimization algorithms for enhanced computational efficiency and tractable inference. Our family of flexible models is meant to provide a practitioner insights regarding the incremental gains in predictive accuracy with enhanced modeling components.
We focused on predicting the amount of time a user might spend on a platform, at the beginning of the user session. 

In our experimental section, we have shown that our method is performing better than the state of the art in this context. Furthermore, we have shown that our model is better for heavy users as well as for users with few sessions. Due to the flexibility of our models, we can achieve lower prediction error for users with many sessions. We can also flexibly incorporate the choice of loss functions that are more robust to outliers in the data.
Our results show that our models can lead to an improvement of up to $4.3$\% in \MAE~on real-life data. Some of our models can be 
$22$ to $43$ times faster (with a $1$ to $2$\% improvement in \MAE) in training time; and $30$ to $500$ times faster in prediction time.

So far we have always investigated a scenario where the model is learned off-line, and it tries to predict the user session length with only few static and off-line extractable features. In the future we aim to extend our model to an on-line version. Furthermore we want to investigate the session length prediction utility within advertising or recommender systems context.

\bibliographystyle{./style/ACM-Reference-Format}

\bibliography{paper}
\end{document}

\end{document}